\documentclass[runningheads]{llncs}

\usepackage[T2A]{fontenc}
\usepackage[utf8]{inputenc}
\usepackage{graphicx}
\usepackage{verbatim}
\usepackage{xcolor}
\begin{document}

\title{Russian Web Tables: A Public Corpus of Web Tables for Russian Language Based on Wikipedia\thanks{This paper was accepted to the XXIV International Conference on Data Analytics and Management in Data Intensive Domains (DAMDID'22).}}

\titlerunning{Russian Web Tables: A Public Corpus of Web Tables for Russian Language}
%
\author{Platon Fedorov\inst{1}\orcidID{0000-0001-8838-6849} \and
Alexey Mironov\inst{1}\orcidID{0000-0003-1828-4606} \and
George Chernishev\inst{1,2}\orcidID{0000-0002-4265-9642}}

\authorrunning{P. Fedorov et al.}

%
\institute{Unidata, Russia \and
Saint-Petersburg State University, Russia\\
\email{\{platon.fedorov, alexey.mironov, georgii.chernyshev\}@unidata-platform.ru}}
\maketitle              
\begin{abstract}

Corpora that contain tabular data such as WebTables are a vital resource for the academic community. Essentially, they are the backbone of any modern research in information management. They are used for various tasks of data extraction, knowledge base construction, question answering, column semantic type detection and many other. Such corpora are useful not only as a source of data, but also as a base for building test datasets. So far, there were no such corpora for the Russian language and this seriously hindered research in the aforementioned areas.

In this paper, we present the first corpus of Web tables created specifically out of Russian language material. It was built via a special toolkit we have developed to crawl the Russian Wikipedia. Both the corpus and the toolkit are open-source and publicly available. Finally, we present a short study that describes Russian Wikipedia tables and their statistics.


\end{abstract}

\graphicspath{./figures}



\section{Introduction}

Tabular data is very important for the information management community since tables are a core data stucture which both academics and practicioners utilize in their work. 

The first studies concerning tabular data appeared almost twenty years ago~\cite{10.1007/3-540-45869-7-29}, and since then there was a constant demand for open tabular data. Table-related studies needed to ensure repeatability and this secured the demand for table corpora, which researchers addressed using the tables available on the Web. 

Currently, Web tables are the building blocks of many modern research in information management. Numerous studies have used table corpora for various data management tasks~\cite{10.1145/3372117}. For example, four major corpus papers~\cite{10.1007/978-3-319-25007-6-25,DBLP:conf/webdb/CafarellaHZWW08,7406328,10.1145/2872518.2889386} have over a thousand of citations combined, according to Google Scholar.

There are several such corpora for the English language~\cite{10.1145/3372117}: they are usually based on pre-crawled data. However, there are no public dedicated table corpora for Russian language that we are aware of.

In this paper, we present the first corpus of Web tables (named RWT, Russian WebTables) created specifically out of Russian language material. For this purpose, we have created a special toolkit that processes Russian Wikipedia. 

Designing this toolkit, we were driven by the following considerations:
\begin{enumerate}
    \item Light-weightness. Existing toolkits for corpus creation usually require a cloud environment just to deploy the initial data. We aim to provide for low-budget research projects and eventually we hope to make corpus creation accessible even to individual students.
    \item Full cycle. Existing toolkits rely on pre-crawled data. Our idea is to create a full cycle system that will not depend on external data and will allow better temporal granularity control. 
    \item Customizability. Our goal is to provide users with means of managing the collection process, for example, allow them to filter tables that contain no latin characters.
\end{enumerate}

It is implemented in Python, has a modern codebase, contains a minimal set of dependencies, and is equipped with a GUI.

Moreover, in this paper we describe the collected corpus: first, we characterize it in terms of high-level metrics, then we outline statistics related to table contents, and finally, we highlight interesting tables and pages.

Overall, the contributions of this paper are the following:

\begin{itemize}
    \item The first corpus of web tables for Russian language, created from Russian Wikipedia. 
    \item A configurable, light-weight, full cycle toolkit for creating such corpora, intended for low-budget projects.
    \item A study that describes tables of the Russian Wikipedia and presents their statistics.
\end{itemize}

Both corpus\footnote{\url{https://gitlab.com/unidata-labs/ru-wiki-tables-dataset}} and toolkit\footnote{\url{https://gitlab.com/unidata-labs/ru-wiki-tables-backend} \url{https://gitlab.com/unidata-labs/ru-wiki-tables-frontend}} are open-source and publicly available.

\section{Background, Motivation, and Related Work}

There are many types of Web tables and their classifications~\cite{10.1145/3372117}. One of the most popular is the following:
\begin{enumerate}
    \item Layout: navigational and formatting. The former are used for navigating within a website and the latter are used for formatting purposes. 
    \item Content: relational, entity, and matrix. Entity tables describe a single entity, relational tables describe a set of entities and their attributes, and matrix-type tables are simply three dimensional datasets.
\end{enumerate}

Relational tables are of the most interest since they contain usable data. However, only a small subset of tables is relational.

Web tables are extensively used for many tasks, namely~\cite{10.1145/2872518.2889386,10.1145/3372117}:

\begin{enumerate}
    \item Table type identification, for example, according to the aforementioned classification.
    \item Table interpretation, which is one of the following: 1) column type identification, 2) entity linking, and 3) relation extraction.
    \item Table search, which can be either keyword-based or table-based search. Recently, there was a surge of interest in this task, related to dataset exploration problems~\cite{DBLP:journals/pvldb/RezigBFPVGS21,DBLP:journals/pvldb/CasteloRSBCF21}.
    \item Knowledge base augmentation and construction.
    \item Table augmentation, which can be done either by extending by row, column, or by doing data completion.
\end{enumerate}

All these tasks require corpora to ensure repeatability of approaches, methods, and algorithms. Moreover, with the popularization of machine learning approaches, tables as training data are in demand too. Therefore, the last decade experienced a boom of table corpus creation.

Unfortunately, this boom has bypassed the Russian database community, and, consequently, Russian language~--- all existing corpora concern either English language exclusively or ignore the language issue altogether. Therefore we have decided to create the RWT corpus~--- a collection of Web tables containing Russian language material. As the starting point we have selected the Russian part of Wikipedia. 

To crawl and process it we have developed a special toolkit~--- the RWT toolkit. The following considerations were taken into account during its creation: \textbf{light-weightness}, \textbf{full cycle support}, and \textbf{customizability}.

Existing corpora, such as the WDC Web Table Corpus~\cite{10.1145/2872518.2889386} or the Dresden Web Table Corpus~\cite{7406328}, are usualy created from the Common Crawl data\footnote{http://commoncrawl.org/}. It is a huge repository of web pages that are crawled monthly and made available to the general public in a compressed form.

For example, the latest version (June/July 2021) contains more than 100TB of compressed data. It will be difficult even to unpack such amount of data on a single machine.

Therefore, relying on the Common Crawl dataset would require a multi-machine environment. For example, in case of the WDC Web Table Corpus, the Amazon EC2 cloud services are needed\footnote{http://webdatacommons.org/framework/} to perform table extraction (and the setup is built into the source code). Therefore, it is prohibitevely expensive for low-budged projects or individual students to use the Common Crawl dataset. 

Another drawback of using pre-crawled data is the temporal granularity restriction: new versions of Common Crawl are released every month or two. It will not be possible to obtain data every week to monitor the changes on-line. At the same time, changes in tabular data is a very promising novel direction, see, for example, the recent series of works concerning the Janus Project~\cite{9458804,seadata} where changes to the Wikipedia tables are explored.

Another Wikipedia-based corpus, the WikiTables corpus~\cite{10.1007/978-3-319-25007-6-25}, was created by a team from Northwestern University. They propose a corpus and a toolkit that are the closest to ours conceptually. The corpus itself is lightweight (1.6GB in size, containing 1.4 million tables), and the toolkit provides a crawler which can be run on a single machine. However, both the corpus and the toolkit have drawbacks. Considering the corpus,
\begin{itemize}
    \item It is heavily focused on the English language. Only 197 of its tables contain more than 5\% of Cyrillic characters.
    \item It was created in 2015 and not updated since; the reference~\cite{seadata} shows that more than a third of the current number of tables has been created since then.
\end{itemize}

The toolkit issues are also related to its age:
\begin{itemize}
    \item Its code has not been maintained for seven years, and it is impossible to run now. It is full of out-of-date dependencies, which are nearly impossible to satisfy. For some dependencies, it is possible to use newer versions, but significant effort is required to adapt them. However, there are some dependencies which are completely lost, in particular, wigLibrary-2.0 (even googling it yields no results). These ones will require full reimplementation.
    \item The toolkit is implemented in Java, while nowadays most data science-related tasks are performed in Python. A Python-based application is more suitable for integration with contemporary data science projects, as the resulting application will have less dependencies. 
    \item Finally, the code for Wikitables Corpus is a part of a larger project, which involves many side dependencies. For example, it uses a MySQL database.
\end{itemize}

As the result, it takes significant work to extract the necessary functionality, comparable to constructing such a tool from scratch, given the fact that the code is impossible to compile.

The final consideration that we took into account is the need to allow users to customize the corpus. Users can have different needs in relation to corpus contents. For example, they may need to keep only the tables that contain not less than a specified ratio of Cyrillic to Latin characters. Another need that may arise is postprocessing tables in such a manner that columns containing numerical data, Web links, and so on will be discarded.

Given the aforementioned considerations and the fact that we are starting from scratch, we have adopted an alternative approach:

\begin{itemize}
    \item The RWT toolkit is a \textbf{light-weight} pipeline that does not require cloud solutions and can be deployed on a home PC. Due to the size of Russian Wikipedia, it is possible to finish the job in a reasonable time. However, if one wishes to speed up the process we provide users with the ``poor man's approach'' to parallelizing it over several PCs (see the next section for details).
    \item The RWT toolkit is a \textbf{full cycle} pipeline that does not depend on any pre-crawled data: it is a crawler and extractor rolled into one.
    \item The RWT toolkit allows user to \textbf{customize} the table extraction process and filter out the irrelevant data. A rich Web-based UI provides easy tuning of the pipeline according to user demands.
\end{itemize}

It is implemented in Python, has a modern codebase, contains a minimal set of dependencies, and is equipped with a GUI.

\section{Toolkit}

In this section we describe the Russian WebTables (RWT) toolkit.

\subsection{Workflow, Architecture, and Implementation}

\begin{figure}
    \centering
    \includegraphics[height=0.45\textwidth]{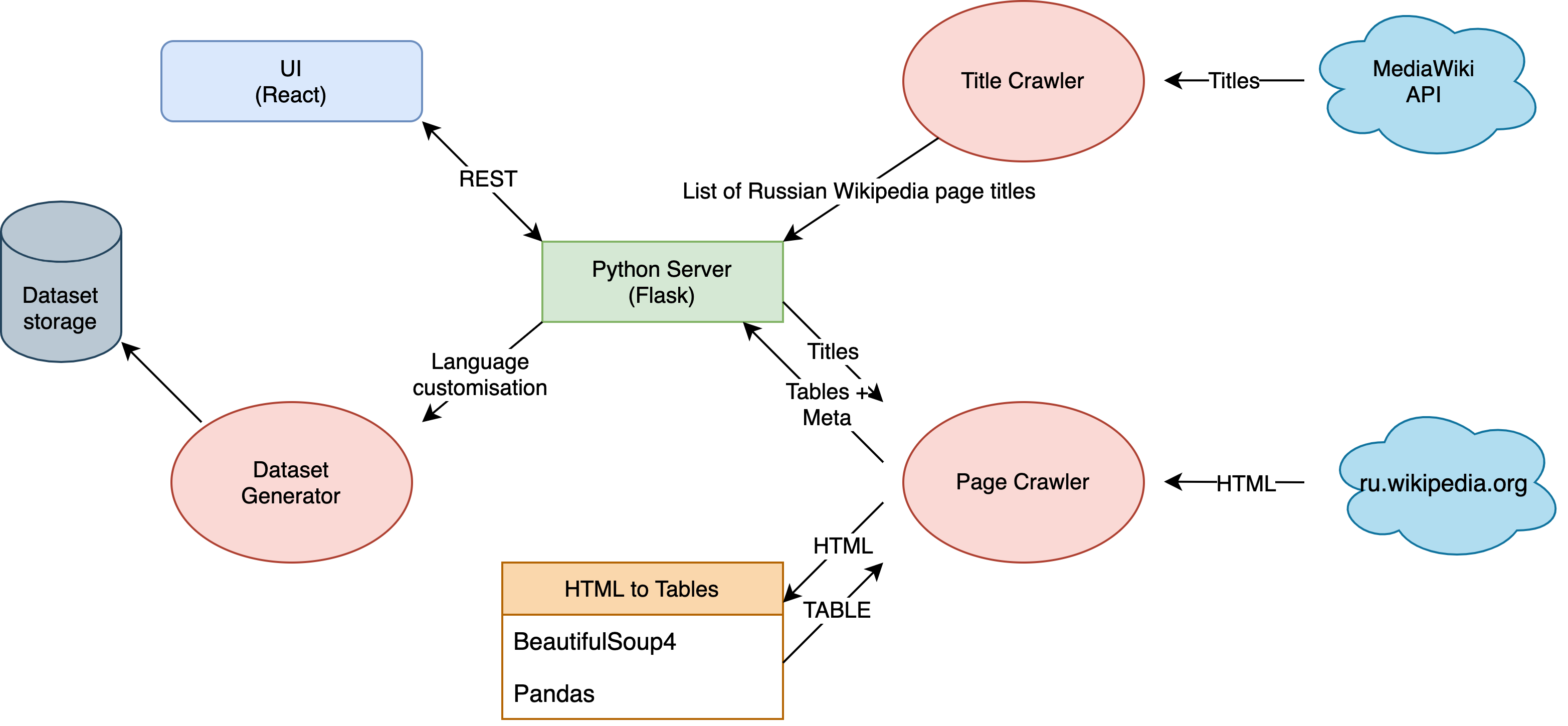}
    \caption{The architecture of the RWT toolkit}
    \label{fig:arch}
\end{figure}

First of all, we will discuss its architecture which is presented in Figure~\ref{fig:arch}. Our solution is built using Python with the following technical stack: the Flask web server, React SPA, BeautifulSoup4, Pandas, and REST API.

The central part of our application is the Flask-based web server which has its UI implemented with the React framework. Its three core modules are:
\begin{enumerate}
    \item Title Crawler, which collects the list of page titles. It uses the MediaWiki API to request titles from Wikipedia.
    \item Page Crawler, which uses these page titles to request individual HTML pages. Later they are fed into BeautifulSoup4 and Pandas to extract their structure and to find tables.
    \item Dataset Generator, which performs language customization for a given table according to user requirements. The resulting table is placed into the Dataset storage.
\end{enumerate}

\begin{figure}
    \centering
    \includegraphics[height=0.9\textwidth,angle=-90]{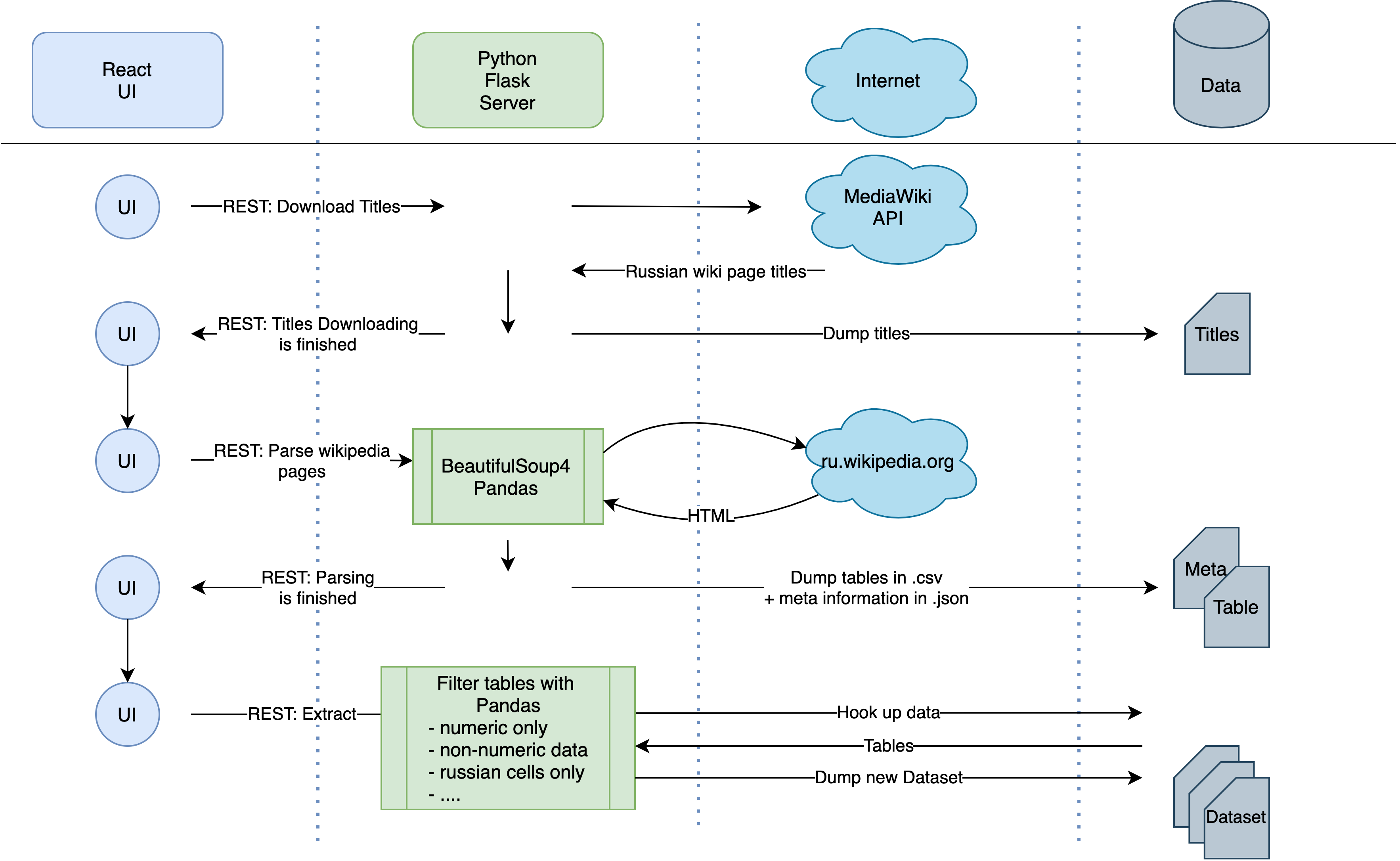}
    \caption{The workflow of the RWT toolkit}
    \label{fig:workflow}
\end{figure}

Next, the workflow of our application is described in Figure~\ref{fig:workflow}. It is as follows:

\begin{enumerate}
    \item At the start of corpus creation, server requests a list of all available page titles for a date, which is specified by the user.
    \item Then, pages are dumped into the local storage and can be divided into parts for parallelization purposes (see the next section for details).
    \item These titles or their part are used to retrieve individual pages. 
    \item Each page is processed with BeautifulSoup4 that allows to extract tables easily.
    \item Then, table metadata is extracted and stored in the JSON format.
    \item Finally, the table is fed into Pandas to perform language customization: row, column, and table transformation and filtering. The resulting tables are stored in the CSV format and are ready for querying.
\end{enumerate}

One must note, that we tried to make our tool resilient to various connection issues. The RWT toolkit is capable of automatically resuming download after such interrupts.

\subsection{RWT toolkit features}\label{sec:features}

\begin{figure}
    \centering
    \includegraphics[height=0.5\textwidth]{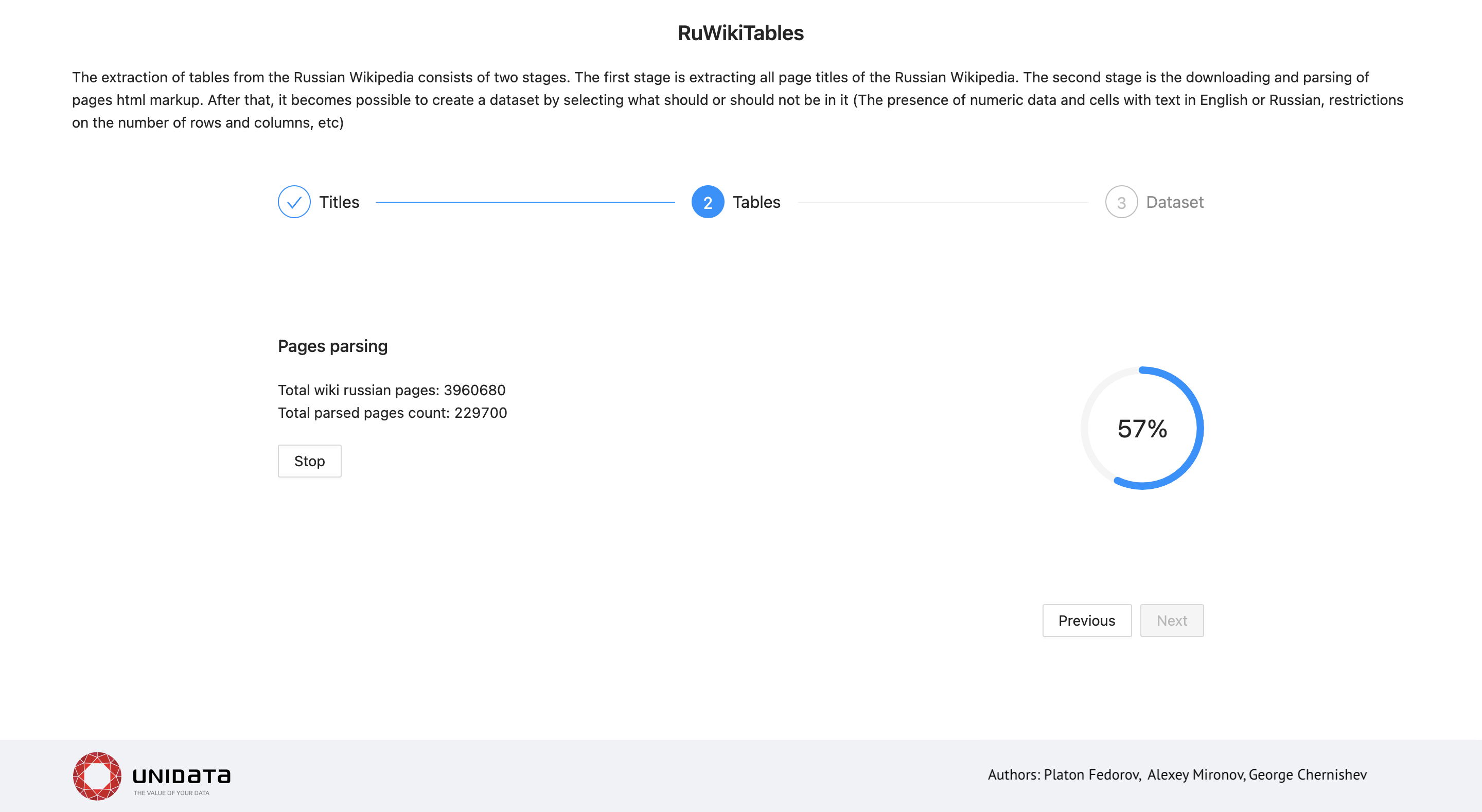}
    \caption{The UI of the RWT toolkit}
    \label{fig:ui}
\end{figure}

The overall feature list of the RWT toolkit is as follows:
\begin{enumerate}
    \item Russian language customization. A user can customize table selection process by employing various filters such as the minimal ratio of Cyrillic characters in the table, or whether to drop columns containing only Latin characters or not and so on.
    
    \item Progress bar. The toolkit offers precise details on how much work is left to perform, thus allowing the user to reliably estimate how much time it will take to finish the job. The details are provided in terms of the number of pages left to process and an average time per page.
    
    \item Pause button. A user can pause and resume download anytime they wish. This feature allows to construct corpus during periods of user inactivity (e.g. at night) or low network utilization, thus allowing to minimize impact on everyday tasks, that are likely run on a household PC.
    
    \item Parallelization. A special kind of download parallelization technique is provided. After fixing a date, the user may select a number of chunks and the chunk to download. Then, the toolkit will download and process pages belonging to the selected chunk. This way, the user can parallelize the download by running the process on their desktop and laptop computers and then unifying the results by copying the corresponding folders. The toolkit is distributed as a Docker container.
    
    \item Our experiments demonstrated (see Table~\ref{tbl:stats}) that constructing a corpus by crawling Wikipedia requires about 70 hours on a single machine. While in our opinion this is a reasonable requirement, some users may find it excessive. For this reason, we have added an option for using Wikipedia dumps\footnote{https://dumps.wikimedia.org/backup-index.html}.
    
    However, using dumps has its own drawbacks:
    \begin{itemize}
        \item First of all, dumps are created irregularly (``These snapshots are provided at the very least monthly and usually twice a month''). If a user needs the freshest data (or data at a particular time), using a dump is not an option.
        \item Only the latest dump is available at a given time. If a user needs a previous version of some page, the dump will not contain it.
        \item Dumps may be discontinued at any given time.
    \end{itemize}

    \item Querying the corpus. A rich number of search options, such as search by title, number of rows or columns, and other metadata is provided. This search is performed over the stored corpus and thus does not require intenet connection.
\end{enumerate}

Concluding this section, we would like to point out the toolkit's modern user interface implemented using the Ant Design React library. An example is presented in Figure~\ref{fig:ui}.

\begin{figure}
    \centering
    \includegraphics[width=0.8\textwidth]{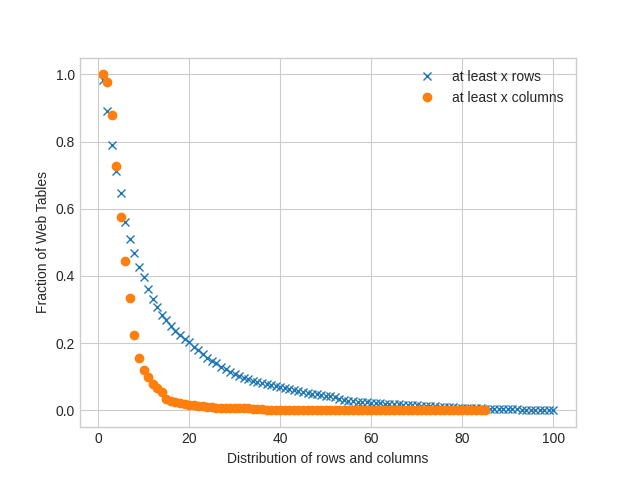}
    \caption{Distribution of \# rows \& columns}
    \label{fig:distrib}
\end{figure}

\subsection{Metadata}~\label{sec:metadata}

Finally, we would like to discuss what kind of metadata we save, inspired by the WDC Web Table Corpus~\cite{10.1145/2872518.2889386}. Similarly to this work, for each table we store its URL, page title, table title, and 200 words before and after the table respecting section borders. Additionally, we store the number of rows, columns, and whether they are numeric.

Since there are no built-in identifiers for tables in Wikipedia itself, we had to create a surrogate one. It consists of the page id (stable over different versions of extracted data) and an offset~--- a position among all tables in this page. Unfortunately, a single offset is insufficient since a new table may be inserted in the middle of document and all subsequent ones will shift. Thus, an additional table matching procedure will be required if it would be necessary to track table evolution.

\section{Tables of Russian Wikipedia}

Apart from creating the toolkit, we have also explored the extracted corpus and provide collected statistics. For this we had not employed any language customization filters and therefore provide vanilla data. We used a Wikipedia snapshot taken on September 13, 2021 (data was crawled and no dumps were used).

First of all, we plot the distribution of columns and rows over the whole corpus, which is depicted in Fig.~\ref{fig:distrib}. This distribution is similar to the one from the study~\cite{10.1145/2872518.2889386}.

Next, we also present the statistics of the entire corpus (Table~\ref{tbl:stats}) in two parts:
\begin{enumerate}
    \item a high-level overview (top part of the table), e.g. number of tables, pages, size, etc;
    \item an in-depth statistics related to table contents (bottom part of the table).
\end{enumerate}

Most of the contents of these tables are self-explanatory, but some comments are need for the following:
\begin{itemize}
    \item ``Rows that are mostly NULL''~--- rows that have more than 70\% empty cells;
    \item ``Columns that are mostly NULL''~--- columns that have more than 70\% empty cells;
    \item ``Construction time in a single PC scenario''~--- here we provide a number that we obtained experimentally on a single home PC with medium-level characteristics;
    \item ``Percentage of cells with non-string data''~--- the percentage of cells that contain at least one non alphabetical character;
\end{itemize}

\begin{table}[]
\centering
\caption{Data statistics of the entire corpus}
\label{tbl:stats}
\begin{tabular}{l|r}
\hline
Statistic & Value \\ \hline 
Total number of pages  & 3960680 \\ \hline
Total number of tables  & 1266731 \\ \hline
Total number of rows  & 15879688 \\ \hline
Total number of columns & 7419771 \\ \hline
Total number of cells & 99638194 \\ \hline
Avg number of cells per table & 81.78\\ \hline
Avg number of tables per page & 0.31 \\ \hline
Construction time in a single PC scenario & $\approx$ 70 hours \\ \hline
Total corpus size (uncompressed)  & $\approx$ 17 GBs \\ \hline \hline \hline

Avg number of cells per row  &  6.27 \\ \hline
Avg number of cells per column &  13.42     \\ \hline
Avg number of characters per cell & 17.49 \\ \hline
Avg number of Cyrillic characters per cell & 11.44 \\ \hline
Avg number of Latin characters per cell & 1.64\\ \hline
Percentage of cells with non-string data & 73\% \\ \hline
Percentage of rows that are mostly NULL & 2\% \\ \hline
Percentage of columns that are mostly NULL & 6\%\\ \hline
Percentage of columns that contain only Cyrillic characters & 3\% \\ \hline
Percentage of columns that contain only Latin characters & 2\%\\ \hline
Percentage of columns that contain only numeric data & 17\%\\ \hline
\end{tabular}
\end{table}

\begin{table}[!htb]
    \begin{minipage}{.5\linewidth}
\centering
\caption{Ten most frequent table sizes}
\label{tbl:sizes}
\begin{tabular}{|l|l|}
\hline
Size & Frequency \\ \hline
3x5  & 41244 \\ \hline
6x2  & 24962 \\ \hline
2x4  & 19538 \\ \hline
7x1  & 17798 \\ \hline
3x4  & 16242 \\ \hline
7x2  & 14917 \\ \hline
3x2  & 14773 \\ \hline
2x1  & 14310 \\ \hline
2x3  & 13556 \\ \hline
4x3  & 13388 \\ \hline
\end{tabular}
    \end{minipage}%
    \begin{minipage}{.5\linewidth}
      \centering
\caption{Ten most common headers}
\label{tbl:headers}
\begin{tabular}{|l|l|}
\hline
Size & Frequency \\ \hline
253717 & Год \\ \hline
137716 & Место \\ \hline
88378 & Название \\ \hline
88197 & Голы \\ \hline
80904 & Роль \\ \hline
79875 & Команда \\ \hline
77080 & Результат \\ \hline
72987 & Сезон \\ \hline
70004 & Игры \\ \hline
68074 & Результаты \\ \hline

\end{tabular}
    \end{minipage} 
\end{table}

Let us discuss Table~\ref{tbl:stats} first. Here we can see the following:
\begin{enumerate}
    \item The share of columns that contain only Cyrillic characters is around 3\%. Therefore, our language customization module is essential, if the goal is to build a ``pure'' textual corpus. Interestingly, the share is comparable with the Latin character share.
    \item The share of pure numeric columns is 17\%, which also indicates the need for filtering, for example, if a text-processing machine learning algorithm would be run on the data.
    \item It would probably be useful to not just prune individual columns, but also post-process contents of individual cells. This is corroborated by the following  facts: the share of non-string data being 73\% and low numbers of ``pure'' Cyrillic columns. This may happen due to links, footnotes, etc.

\begin{table}[]
\centering
\caption{Interesting tables}
\label{tbl:facts}
\begin{tabular}{|l|p{7cm}|}
\hline
Statistic & Page Title \\ \hline
Widest table (47)  & Чемпионат мира по ралли \\ \hline
Longest table (4098 cells)  & Области для частного использования \\ \hline
Most populated table (680827 characters)  & Список кандидатов по одномандатным округам на выборах в Государственную думу 2021 года \\ \hline
Most populated table (69666 cells)  & Области для частного использования \\ \hline
\end{tabular}
\end{table}

\begin{table}[]
\centering
\caption{Top 10 most table-rich pages}
\label{tbl:rich}
\begin{tabular}{|l|p{11cm}|}
\hline
Tables & Title \\ \hline
222 & Список кавалеров ордена Святого Духа \\ \hline
192 & Чемпионат мира по кёрлингу среди смешанных пар 2019 \\ \hline
190 & Футбольные клубы Латвии в еврокубках \\ \hline 
175 & Чемпионат мира по кёрлингу среди смешанных пар 2018 \\ \hline
173 & Список министров промышленности России \\ \hline
166 & Список видео консолей первого поколения \\ \hline
156 & Деревянная архитектура Нижнего Новгорода \\ \hline
136 & Рекорды основного тура ATP \\ \hline
132 & Чемпионат Европы по кёрлингу 2019 \\ \hline
127 & Список бомбардиров Кубка Лиги чемпионов по странам \\ \hline
125 & Список операций вооружённых сил СССР во Второй мировой войне \\ \hline

\end{tabular}
\end{table}

\begin{table}[]
\centering
\caption{Top 10 most table-rich pages (filtered: $rows \ge 3$, $columns \ge 5$)}
\label{tbl:rich-filtered}
\begin{tabular}{|l|p{11cm}|}
\hline
Tables & Title \\ \hline
96 & Прыжок в высоту (лучшие спортсмены года, женщины) \\ \hline
96 & Список бомбардиров Кубка Лиги чемпионов по странам \\ \hline
79 & Список видео консолей первого поколения \\ \hline
68 & Бразилия на летних Олимпийских играх 2016 \\ \hline
68 & США на летних Олимпийских играх 2016  \\ \hline
60 & Первый дивизион чемпионата мира по хоккею с шайбой 2015 (составы) \\ \hline
60 & Парламентские выборы в Молдавии (2019) \\ \hline
58 & Искатель (альманах) \\ \hline
58 & Россия на летней Универсиаде 2013 \\ \hline
50 & Франция на летних Олимпийских играх 2016 \\ \hline
50 & Орёл и решка (телепередача) \\ \hline

\end{tabular}
\end{table}

    \item There is a noticeable amount of rows and columns that have a lot of empty cells (2\% and 6\%).
\end{enumerate}

Tables~\ref{tbl:sizes} and~\ref{tbl:headers} shows the most common table sizes and most common table headers that we have encountered. Interestingly, these are mostly small tables which are usually belong to the entity type. Regarding the headers we have have filtered out the ``trivial'' header names. Trivial include, for example, digits like ``0'', ``1'', and so on. In the original version of the ranking list the digits from 0 to 7 were present. Since they are not of any value for data scientists we have decided to remove them. Finally, we have noticed that the distribution of these digits looks rather similar to Benford's law\footnote{\url{https://en.wikipedia.org/wiki/Benford's\_law}}.

Table~\ref{tbl:facts} presents interesting facts~--- notable tables and pages. Additionally, we have listed most table-rich pages in Table~\ref{tbl:rich}. Both this table and Table~\ref{tbl:headers} demonstrate the dominance of tables related to sports in Russian Wikipedia. Unfortunately, sports-related tables provide little to no value for information management problems, since they are usually short~--- spanning one or two rows, and containing mostly numbers (points, number of wins, goals scored, etc) and names. Interstingly, these tables also produce the headers with single-digit names discussed earlier. We tried to eliminate them by filtering tables by size (Table~\ref{tbl:rich-filtered}). The resulting set contains more meaningful tables, but still, sports-related tables prevail in it. Therefore, in order to eliminate this kind of tables completely, advanced methods of filtering (such as machine learning) are needed.

\section{Acknowledgments}

First of all, we would like to thank anonymous reviewers for their helpful feedback. Next, we would like to thank Dmitrij Koznov for his remarks and suggestions which arose during our discussions. Finally, we also would like to thank Anna Smirnova for her help with the preparation of the paper.

\section{Conclusion and Future Work}~\label{sec:conl-fw}

This paper presents the first corpus of Web tables created specifically out of Russian language material. To the best of our knowledge, this is the only such corpus currently publicly available outside large search engine companies.

We also describe the toolkit that we have developed for this task. It is a customizable and light-weight pipeline that is oriented towards low-budget research and therefore, does not require cloud solutions and can be deployed on home PCs. Its customizability allows users to run a fine-grained table selection during the creation of corpus.  Finally, our toolkit allows both to crawl data to obtain the freshest snapshot and to use Wikipedia dumps as a faster option.

In the future we plan to expand the corpus with open governmental data, from resources such as \url{https://data.gov.ru/} and \url{https://data.gov.spb.ru/}. They offer collections of large tables that contain diverse information related to many different domains: art, education, infrastructure, commerce, etc.

We also intend to develop the toolkit further, and our first goal will be to expand the functionality of our tool. In particular, we would like to enable the crawler to collect data for a given date in order to scrape previous versions of pages and their tables. Other planned improvements are the following: table type detection (e.g. relational or entity), a module for checking joinability and unionability~\cite{DBLP:journals/pvldb/CasteloRSBCF21,DBLP:journals/pvldb/OuelletteSNBZPM21}, semantic type detection of columns~\cite{10.5555/3430915.3442430,10.1145/3292500.3330993}, and many more.

\bibliographystyle{splncs04}
\bibliography{mybib}

\end{document}